\documentclass[a4paper]{article}

\usepackage{itgspeech2018}    %
\usepackage{times}            %
\usepackage[english]{babel}   %
\usepackage[utf8]{inputenc}%
\usepackage[T1]{fontenc}      %
\usepackage[sort&compress,numbers]{natbib}	%
\usepackage{amsmath,amssymb}
\usepackage{graphicx}
\usepackage[colorlinks=false,pdfborder={0 0 0}]{hyperref}
\usepackage{units}

\usepackage{booktabs}

\usepackage{microtype}

\title{Open Source Automatic Speech Recognition for German}

\author{Benjamin Milde$^1$, Arne Köhn$^2$}

\address{Language Technology$^1$ and Natural language Systems$^2$ group, FB Informatik, Universität Hamburg, Germany\\
  Email: \texttt{\{milde, koehn\}@informatik.uni-hamburg.de.de}}

\begin{document}

\maketitle

\begin{abstract}

  High quality Automatic Speech Recognition (ASR) is a prerequisite
  for speech-based applications and research. While state-of-the-art
  ASR software is freely available, the language dependent acoustic
  models are lacking for languages other than English, due to the
  limited amount of freely available training data.
  We train acoustic models for German with Kaldi on two datasets, which are both
  distributed under a Creative Commons license. The resulting model
  is freely redistributable, lowering the cost of entry for German
  ASR. The models are trained on a total of 412 hours of German read speech data and we achieve a relative word error reduction of 26\% by adding data from the Spoken Wikipedia Corpus to the previously best
  freely available German acoustic model recipe and dataset. Our best model achieves a word error rate of 14.38 on the Tuda-De test set. 
  Due to the large amount of speakers and the diversity of topics
  included in the training data, our model is robust against speaker
  variation and topic shift.
  
\end{abstract}

\section{Introduction}
\label{sec:introduction}

Over the past years a lot of progress has been made to make Automatic Speech Recognition (ASR) more robust and practicable, mainly due to incoporating (deep) neural networks as central part of the recognition pipeline. There has also been a shift towards making the underlying recognition software more accessible. With the introduction of the Kaldi toolkit \cite{povey2011kaldi}, a state-of-the-art open source toolkit for speaker-independent large vocabulary ASR became available for researchers and developers alike. Over the past years, it has evolved into a very popular open source ASR toolkit, either pushing state-of-the-art acoustic models or following their performance closely.

For English, open source resources to train Kaldi acoustic models as well as language models and a phoneme lexicon exist, in sufficient quality and quantity: TED-LIUM \cite{rousseau2014enhancing, hernandez2018ted} and Librispeech (1000h) \cite{panayotov2015librispeech} allow large-scale training of speech recognizers, with word error rates in the low single digits in their respective domains (6.5\% WER for presentation speech, 3.2\% WER for clean read speech  \cite{han2017capio}). These resources exist alongside proprietary resources, such as: TIMIT \cite{garofolo1993darpa}, Switchboard \cite{godfrey1992switchboard} and the Fisher corpus \cite{cieri2004fisher}. The latter two also enable low word error rates (WERs) on more difficult spontaneous conversational telephone speech test sets (e.g. for Switchboard 5.5\% WER in \cite{saon2017english}, within close range of human performance).

However, models trained from open source and freely available resources allow personal, academic and commercial use cases without licensing issues, lowering the barrier of entry.  Having access to a locally running speech recognition software (or a private server instance) solves privacy issues of speech APIs from cloud providers. English speech recognition models for Kaldi are available as pretrained packages or freely available training recipes and these models are used in the wild for down-stream NLP applications, e.g. \cite{oualil2017context, milde2016ambient}. We would like to establish the models presented in this paper as go-to models for open source German speech recognition with Kaldi -- with freely available training recipes, making it easily extensible, as well as offering pre-trained models. In the remainder of the paper we discuss the freely available data resources for German and our recognition results.

\begin{figure}
  \centering
  \includegraphics[width=0.95\columnwidth]{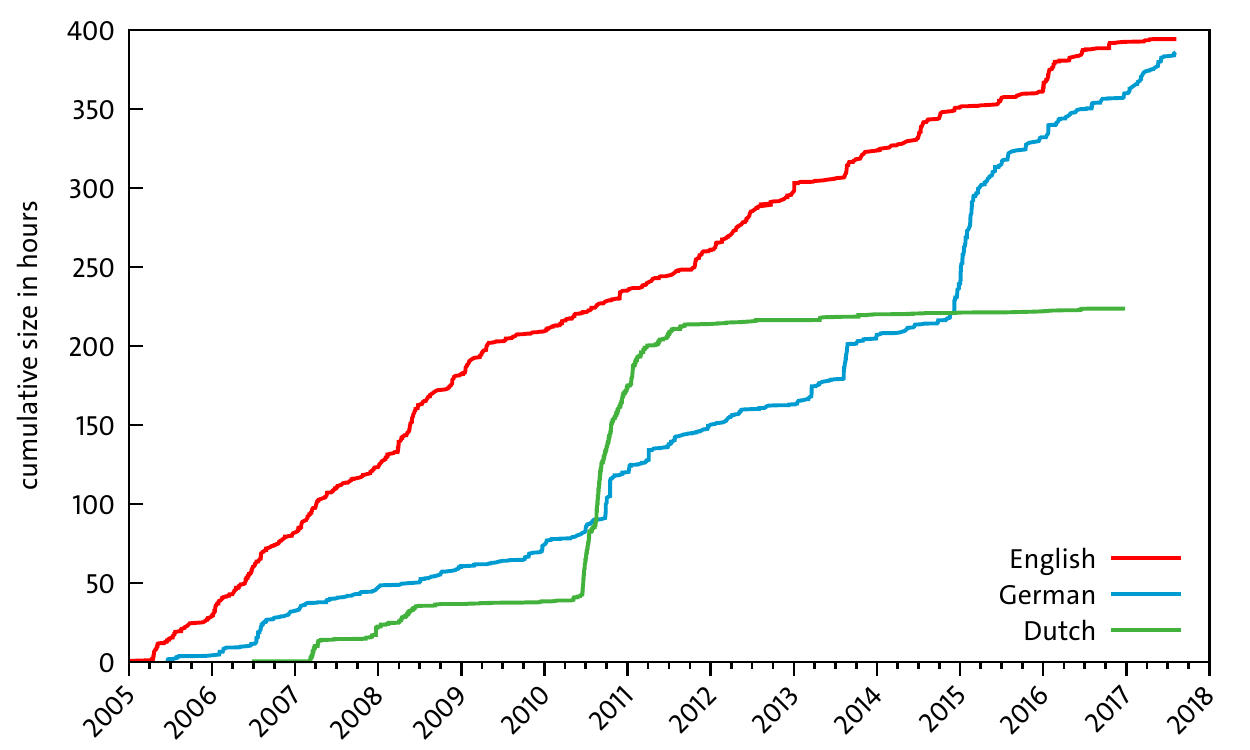}
  \caption{Growth of the English, German and Dutch Spoken Wikipedia resources over time. English and German both grow with about 33h of additional audio per year.}
  \label{fig:swc-over-time}
\end{figure}

One of our data resources is automatically aligned data from the Spoken Wikipedia project. This is a very interesting resource, as new speech data is consistently added to the project by volunteers (see the growth rate in Figure~\ref{fig:swc-over-time}) and the training process can be extended form time to time with new data. Our final model can deal with different microphone and unknown speakers in an open vocabulary setting. 

\section{Data Sets}
\label{sec:data-sets}

In the following, we briefly describe the data resources that we used to train our models. Also, in Table \ref{tab:trainingdatalength} we give an overview of the amount of available training data. All of the following resources are freely available and are published with permissive Creative Commons open source licenses (i.e. free for all, commercial use allowed).

\begin{table}[t]
  \centering
  \begin{tabular}{lcc}
    \toprule
\textbf{Dataset} &    \textbf{Training hours} &  \textbf{Speakers} \\  
    \midrule
Tuda-De & 127h & 147 \\
SWC German (cons) & 141h & 363 \\
SWC German  (min) & 285h & 363 \\
    \midrule
\textbf{total (cons)} & 268h & \textbf{510} \\
\textbf{total (min)} & \textbf{412h} & \textbf{510} \\
    \bottomrule
  \end{tabular}
  \caption{Amount of training data and speakers from our two open source datasets that we used to train our Kaldi models. cons: conservative pruning, min: minimal pruning}
  \label{tab:trainingdatalength}
\end{table}

\subsection{Spoken Wikipedia Corpus}
\label{sec:swc}

The Spoken
Wikipedia\footnote{\url{https://en.wikipedia.org/wiki/Wikipedia:WikiProject_Spoken_Wikipedia}}
is a project run by volunteers to read and record Wikipedia articles.
The audio files produced are linked to the Wikipedia articles,
with semi-structured metadata attached.  Sub-projects exist for many
languages, but English, German, and Dutch are the largest ones, by a
large margin.
The Spoken Wikipedia Corpora \cite{Baumann2018} (SWC) are a collection of
time-aligned spoken Wikipedia articles for Dutch, English and German
using a fully automated pipeline to download, normalize and
align the data.  Crucially, the exact correspondences to original
articles is preserved in the alignment data.  For German, both an
alignment of normalized words as well as a phone-based alignment
exists. We use word-based alignments.

Being based on found data, the alignments are not perfect: Parts of
the articles are not aligned at all, e.g.\ because of incorrect
normalization or pronunciation that deviates from the expectation.
For material such as tables or formulas it is unknown how
they will be read (or if they are read at all) and they are therefore
excluded from the alignment process. 

Being recorded by volunteers reading complete articles, the data fits
very well how a user naturally speaks, arguably better than a
controlled recording in a lab.  The vocabulary is quite large due to
the encyclopedic nature of the articles. 
Topics of the articles are diverse and range from obscure
technical articles like ``Brainfuck'' (an esoteric programming language)
to a description of ``Isar'' (a river).

\begin{figure}
  \centering
  \includegraphics[width=0.95\columnwidth]{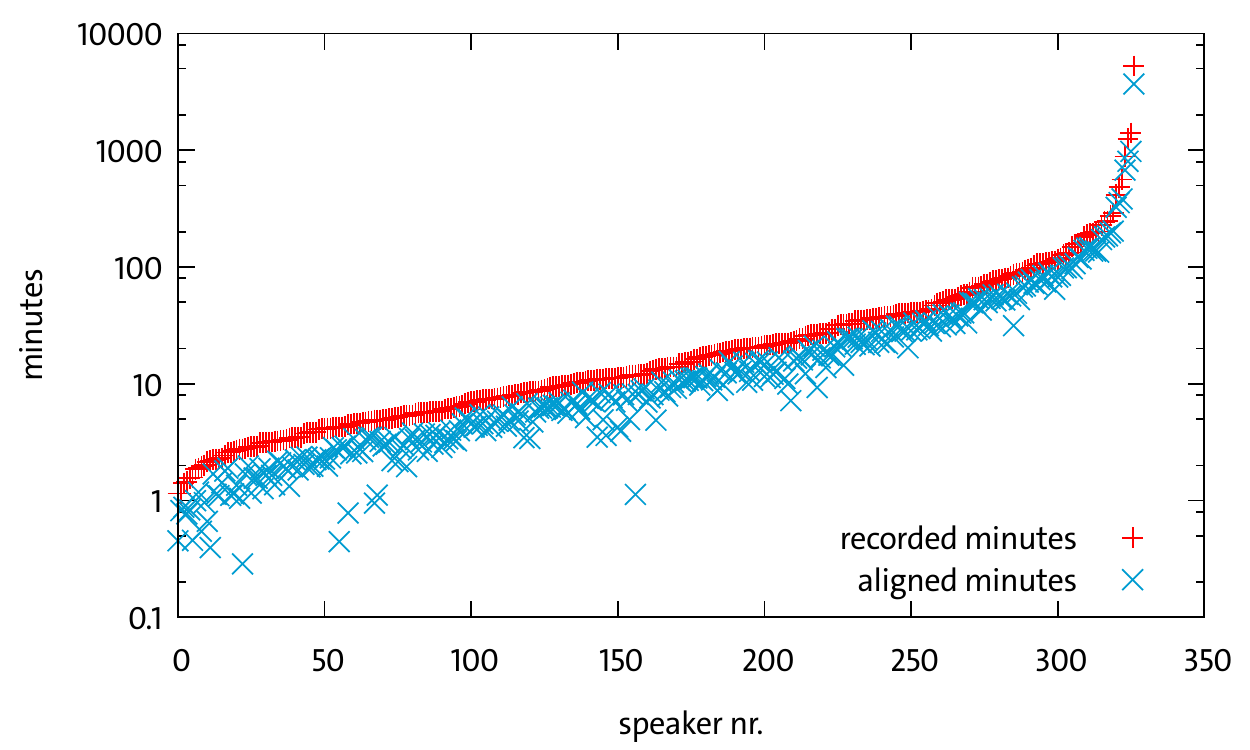}
  \caption{Distribution of speaker contribution and amount of aligned material for the SWC corpus.}
  \label{fig:swc-audio-dist}
\end{figure}

To train Kaldi on the Spoken Wikipedia, we adapted the pre-existing
pipeline from the Spoken Wikipedia Corpora which bootstraps the Sphinx speech
recognizer \cite{sphinx2003} used for the SWC alignment by iteratively
training new models on the data of the previous alignment.
As a speech recognizer can not be trained on
partly aligned long audio (some recordings last several hours), the
SWC pipeline contains a snippet extractor which searches for
continuously aligned data of appropriate length. The snippet
extractor generates training segments along Voice Activity Detection (VAD) boundaries and discards
utterances that are too short (smaller than 0.6 seconds), have more
than 20\% of unaligned data, more than two consecutive unaligned words,
an unaligned word at the beginning or end
or pauses longer than 1.5 seconds.
We call this approach \emph{conservative} pruning, as
it tries to minimize errors in the training data.

In a second setting, we deviate from the pre-existing snippet
extraction and extract all segments defined by VAD boundaries for
which at least 65\% of the words are aligned. There are no other
restrictions, e.\,g.\ the start and end words do not need to be
aligned. We call this approach \emph{minimal} pruning, as it keeps
much more training material than the conservative approach. The 
German Spoken Wikipedia has 363 speakers who committed 349h of
audio to the project, of which 249h could be successfully aligned.
Of these, 141h of audio is extracted with conservative pruning and
285h with minimal pruning.

\subsection{Tuda-De}
\label{sec:tuda-de}

In \cite{radeck2015open}, an Open Source corpus of German utterances was described and publicly released, with a focus on distant speech recognition. Sentences were sourced from different text genres: Wikipedia, parliament speeches, simple command and control prompts. Volunteers, mostly students, read the sentences into four different microphones, placed at a distance of one meter from the speaker. One of these microphones was a Microsoft Kinect. The corpus contains data from the beamformed signal of the Kinect, as well as mixed down single channel raw data from the microphone array (due to driver restrictions the raw multi channel data could not be recorded). Yamaha PSG-01S, a simple USB table microphone and a Samson C01U, a studio microphone, were also used to record audio simultaneously. A further simultaneous recording was made with a
built-in laptop microphone (Realtek), at a different position in the room and next to a very noisy fan. For nearly every utterance the corpus contains five sound files, apart from a few where driver hiccups resulted in fewer recordings. Four of these streams are fairly clean and comprehensible, the recordings from the Realtek microphone next to a noisy fan are very difficult to understand, even for humans. Female speakers make up about 30\%
of the data and most speakers are between 18 and 30 years old. We use version 2 of the corpus.

\subsection{Lexicon}
\label{sec:lexicon}

MARY-TTS \cite{schroder2003german} is an open source Text-To-Speech (TTS) system. It also contains a manually created phoneme dictionary resource for German, containing 26,231 words and their phoneme transcriptions in a dialect of extended SAM-PA BAS \cite{BASSAMPA}. We use Sequitur \cite{bisani2008joint} to train a grapheme-to-phoneme (G2P) model, to be able to add automatically generated entries for out-of-vocabulary (OOV) words to the lexicon as needed.
For the Tuda-De corpus the final lexicon size is 28,131 words; this includes all words from the MARY lexicon and automatically generated entries for all OOV words in the train set. When we combine the Tuda-De transcriptions with the SWC transcriptions, more OOV lexicon entries need to be automatically generated and the final lexicon size grows to 126,794 words using the conservatively pruned SWC data, respectively 182,784 words with minimally pruned SWC data. To measure the effects of an even larger vocabulary size, we also computed the 300,000 most frequent words in the German Wikipedia (April 2018) and generated additional phonetic entries. We merged the vocabulary with the previous lexicon and obtained a larger lexicon containing 350,029 words.

\subsection{Language Models}

\begin{table*}[t]
  \centering
  \begin{tabular}{llllcc}
    \toprule
\textbf{Model} &    \textbf{Dataset} &    \textbf{Vocabulary} &    \textbf{LM} &  \multicolumn{2}{c}{\textbf{WER}} \\ 
               &                   &                   &                   & \textbf{dev} & \textbf{test} \\  
    \midrule
    GMM-HMM
               & Tuda-De &  28,131 & 3-gram KN  & 45.31 & 45.55 \\
               & Tuda-De & 126,794 & '' & 37.47 & 38.34 \\
               & Tuda-De + SWC (cons. pruned) & '' & '' & 29.97 & 31.06 \\
               & Tuda-De + SWC (min. pruned) & '' & '' & 29.79 & 30.99 \\
               & Tuda-De + SWC  (min. pruned) & 182,784 & '' & 26.92 & 28.25 \\ %
               & Tuda-De + SWC  (min. pruned) & 350,029 & 4-gram KN & 24.91 & 25.77 \\
    \midrule   
    TDNN-HMM
               & Tuda-De &  28,131 & 3-gram KN &  35.53 & 36.32 \\
               & Tuda-De  &  126,794 & '' & 28.08  & 28.96 \\
               & Tuda-De + SWC (cons. pruned) & '' & '' & 20.91 & 22.22 \\
               & Tuda-De + SWC (min. pruned) & '' & '' & 20.30 & 21.43 \\
               & Tuda-De + SWC (min. pruned) & 182,784 & '' & 18.39 & 19.60 \\
               & Tuda-De + SWC (min. pruned) & 350,029 & 4-gram KN & 15.32 & 16.49 \\
               & Tuda-De + SWC (min. pruned) & 350,029 & + 2-layer LSTM LM & \textbf{13.14} & \textbf{14.38} \\

    \bottomrule
  \end{tabular}
  \caption{WER results on the Tuda-De dev and test sets. The scores are for decoding combined data from Kinect (Beam and RAW), Samson and Yamaha microphones.}
  \label{tab:mainresults}
\end{table*}

We used the same text sources as in \cite{radeck2015open} and trained similar baseline language models. In particular, we trained 3-gram and 4-gram language models with Kneser-Ney smoothing \cite{kneser1995improved} and different vocabulary sizes on approximately 8 million German sentences. The sentences are selected from similar sources as the spoken data (Wikipedia, Parliament and some crawled sentences). Also, they are already filtered, so that sentences from the development and test sets of the Tuda-De corpus are not included in LM training texts. All sentences were normalized using the frontend of the MARY TTS software \cite{schroder2003german}, similarly to the normalization process of the SWC corpus.
We also use the newly released Kaldi-RNNLM \cite{xu2018neural} to train a recurrent neural network based LM on the same text sources. We use the same parameters as in the Switchboard LSTM 1e example: two stacked LSTM layers with a cell width of 1024.

\section{Experiments and Evaluation}
\label{sec:experiments}

We use Gaussian Mixture Model (GMM) - Hidden Markov Models (HMM) and Time-Delayed Neural Networks (TDNNs) \cite{waibel1990phoneme, peddinti2015time} as acoustic models following the chain-recipe (s5\_r2) of the TED-LIUM corpus \cite{rousseau2014enhancing} example in Kaldi. The TDNNs have a width of 1024 neurons. For GMM-HMM models, we adapted the Kaldi egs for the Switchboard corpus (swbd s5c, model tri4). As input to the TDNN we also use online i-vectors (helping with speaker adaptation, c.f. \cite{saon2013speaker, senior2014improving, miao2015speaker}). 

As the TDNN-HMM chain models are sequence discriminatively trained on the utterances, they are more prone to overfitting and do not cope well with incorrect transcriptions \cite{povey2016purely}. Since the SWC transcriptions are aligned from found data, we expect that some of the transcriptions could be problematic, particularly when we apply only minimal pruning to SWC. We follow the recipe used in the Kaldi TED-LIUM TDNN example and clean the training data by decoding it and removing utterances which do not match their supposed transcriptions. While analyzing the cleaned utterances, we also noted that some of the Tuda-De utterances are wrongly annotated, mostly because of hiccups in the recording software \cite{schnelle2014open} resulting in (completely) wrongly assigned utterance transcriptions. The cleanup removes about 1.6\% of the Tuda-De data and 6.9\% of the combined Tuda-De and conservatively pruned SWC data (268.5h $\rightarrow$ 250h). With minimally pruned SWC data, 8.8\% of the combined training data is removed from 412 hours, resulting in 375 hours of cleaned training data.

We use the dev and test set from the Tuda-De corpus to measure word error rates (WER). The experiments in \cite{radeck2015open} defined a closed vocabulary task with no out-of-vocabulary (OOV) words, as OOV words in test and dev were added to the lexicon. This makes WER rates somewhat lower in comparison, but a bit unrealistic. In Table \ref{tab:mainresults} we show results for a more realistic open domain setting, where the dev and test vocabulary is not known a priori. Using only a 28,131 word vocabulary yields very high WER for GMM-HMM and TDNN-HMM models alike, because of a high OOV rate. Extending the vocabulary to 126,794 words reduces both GMM-HMM and TDNN-HMM WER by about 20\% relative. Adding SWC data to the Tuda-De utterances improves these TDNN-HMM results significantly, even when we use the same vocabulary size. Using a minimal pruning strategy with the SWC data and subsequently relying more on Kaldi's cleaning scripts gives slightly better results: 26\% relative reduction vs. 23.3 \% relative reduction. Finally, we achieve our best WERs when we use a significantly larger vocabulary and a better LM. Our test score with an open domain vocabulary of 350,029 words is 16.49 WER and can be further improved by using lattice rescoring with an LSTM LM to 14.38 WER. This is a significant improvement over the 20.5 WER (without OOVs) reported in \cite{radeck2015open}.

\begin{table}[t]
  \centering
  \begin{tabular}{llcc}
    \toprule
    \textbf{Model} &   \textbf{Microphone} & \multicolumn{2}{c}{\textbf{WER}} \\ 
                   &                       & \textbf{dev} & \textbf{test} \\  
    \midrule
    TDNN-HMM
                   & Kinect-RAW & \textbf{13.82} & \textbf{15.03} \\
                   & Samson & 14.19 & 15.18 \\
                   & Yamaha & 14.84 & 15.77 \\
                   & Kinect-Beam & 19.12 & 20.86 \\
                   & Realtek & 66.46 & 63.41 \\
\bottomrule
  \end{tabular}
  \caption{WER results of models trained on combined data (Tuda-De and SWC) for the different microphones in the Tuda-De dev and train sets. All WER results above are with the lexicon of 350,029 words and without RNNLM rescoring.}
  \label{tab:microphones}
  \vspace{-3mm}

\end{table}

\subsection{OOV Rate}
\label{sec:oov-rate}

Due to the type of text used in our data sets, the number of
unseen words in the test set is quite high, with an OOV rate of 14\% for the lexicon with 28k entries, 8\% using 126k words and 3.2\% using 350k words. The OOV rate poses a lower bound on achievable WER and also explains the large influence of vocabulary size on observed WERs.

The largest problem the ASR model faced during evaluation was
compounding.  As German is a very productive language, compounds
unknown to the language model are quite frequent, even though the
acoustic model is clearly able to recover the information needed.
Because the language model tends to create more tokens than in the
original text when trying to recognize compounds not in the lexicon,
each of such errors is counted as at least two errors:
a substitution and an insertion error.  For example, the word
``nachzumachen'' is recognized as ``nach zu machen'', resulting in
three recognition errors: two insertions and a substitution.
Overall, about a quarter of the errors (2.6k of the 11.5k with the 350k vocabulary model) are part of a sequence of insertions
followed by a substitution, indicative for an error as just described.
We manually checked a sample of these errors and could verify that
indeed most of them are compounds recognized as multiple word such as
``Umweltvereinbarung'', ``Fußpfad'', or ``zweitausendzwölf''.

\subsection{Differences Between Microphones}
\label{sec:diff-betw-micr}

The dev and test set of Tuda-De is recorded using multiple microphones. In Table~\ref{tab:microphones} we calculated WER individually on the dev and test sets per microphone.
The differences in recognition accuracy are surprisingly small for Kinect-RAW, Samson and Yamaha recordings. The usual range of WER for TDNN-HMM models we observed for these microphones is between 15.03\% and 15.77\%. However, the beamformed WER result for the Kinect is significantly higher than decoding the raw (mixed down to one channel) data. The beamforming algorithm of the Microsoft Kinect is closed source, but a few observations are very noticable in the recorded signal. There is a very audible "tin can effect" in the audio signals, probably from a noise suppression algorithm. The beam also seems to get misdirected after pauses, too, c.f. Section 3 in \cite{schnelle2014open}. The recordings were made with automatic gain control, in some of the utterances the beginning is difficult to understand as a result.

An exception to the otherwise good results are also WERs from the Realtek microphone. It produced heavily distorted recordings due to a nearby laptop fan, making these recordings very challenging to decode. The data from this microphone is however not officially part of the dev and test set (it is also not included in Table \ref{tab:mainresults}).

\subsection{Conversational Speech}

In the Verbmobil project (1993-2000), the goal was to establish whether translation of spontaneous speech into other languages is possible \cite{wahlster2013verbmobil}. Conversational speech data was recorded for German, English and Japanese, in the limited domain of scheduling appointments. We used the dev and test data of the first revision of the German subset of the Verbmobil corpus (VM1). Since our acoustic models are trained exclusively on read speech, it provides a good test set showing how well our models cope with a more challenging conversational and spontaneous speaking style.

In Table \ref{tab:vmresults} we show results for decoding VM1 utterances with our acoustic models. We decode with two different vocabularies and FSTs, a general purpose vocabulary (as also used for the results in Table \ref{tab:mainresults}) and a domain specific vocabulary, using the lexicon words of the VM1 corpus (6851 words). For the latter we recomputed our LM with the reduced vocabulary. We do not use the manual lexicon entries of the VM1 corpus and instead use the same lexicon we use in the general purpose case, reducing it and generating automatic OOV phoneme lexicon entries as needed.

The domain specific WER score with limited vocabulary is usually found in the literature for the Verbmobil corpus. A newer reference score for a DNN-HMM trained with Kaldi is 12.5\% WER in \cite{gaida2014comparing}. Our score of 20.04\% WER is probably due not using the optimized and manually generated lexicon as well as due to a mismatch in the training data for the acoustic model (read speech vs. conversational speech). The model in \cite{gaida2014comparing} is exclusively trained on in-domain audio data, while we excluded any proprietary VM1 speech training data and only used our freely available open source speech recordings.

\begin{table}[t]
  \centering
  \begin{tabular}{llcc}
    \toprule
\textbf{Model} & \textbf{Vocabulary} &    \multicolumn{2}{c}{\textbf{WER}} \\ 
              &      & \textbf{dev} & \textbf{test} \\  
    \midrule    
    GMM-HMM & General purp. (\textasciitilde 350k) & 46.42 & 50.56 \\   
    TDNN-HMM & General purp. (\textasciitilde 350k) & 33.69  & 38.23  \\
    \midrule
    GMM-HMM & Domain specific (\textasciitilde 7k) & 27.18 & 29.12 \\   
    TDNN-HMM & Domain specific (\textasciitilde 7k) & \textbf{18.17}  & \textbf{20.04}  \\
    \bottomrule
  \end{tabular}
  \caption{WER results on the Verbmobil (VM1) dev and test data, without RNNLM rescoring.}
  \label{tab:vmresults}
  \vspace{-3mm}
\end{table}

\section{Conclusions and Outlook}
\label{sec:conclusions-outlook}

We have introduced a freely available ASR model for German which
improves upon the previously best one by a large margin, both due to
improvements in algorithms and a significant increase of freely
available data.
The free acoustic model fosters replicable research, and lowers the
cost of entry for (non-cloud based) ASR, as the model can be
readily downloaded\footnote{Training scripts and model files are currently available at: \\ \url{https://github.com/uhh-lt/kaldi-tuda-de/}}. In light of the recent privacy debate on data handling, especially in the EU, freely available acoustic models for German have obvious advantages over cloud based or closed source models. Our models can be run locally and user-recorded speech data does not have to be transferred to a 3rd party cloud provider, where privacy concerns will arise.

Our evaluation shows that the model performs well on new speakers, different microphones with around 14.4\% WER for rescored TDNN-HMM models. The size of the general purporse vocabulary has a large effect on WERs - a large part of the remaining recognition errors are due to vocabulary problems and the underlying language model. We expect a subword unit or decompounding approach to work better than a fixed word approach for German read speech \cite{smit2017improved}. A remaining challenge is conversational speech, but reasonable performance can be achieved with a domain specific vocabulary. On the other hand, as more and more articles are spoken and recorded by volunteers for the Spoken Wikipedia project, we also expect benefits for our acoustic models through the use of the additional data.

The recipes we built for German can also be adapted to other languages. A good candidate, due to data be readily available in the Spoken Wikipedia corpus, is Dutch. The availability of Dutch data outside the Spoken Wikipedia corpus is even more limited than of German data -- there are currently
only ten hours available on Voxforge\footnote{\url{http://www.voxforge.org/nl/Downloads}}, while up to 200 hours can potentially be used for model training from the SWC corpus.

\small
\bibliographystyle{ieeetr}
\bibliography{itg18}

\begin{thebibliography}{10}

\bibitem{povey2011kaldi}
D.~Povey, A.~Ghoshal, G.~Boulianne, L.~Burget, O.~Glembek, N.~Goel,
  M.~Hannemann, P.~Motlicek, Y.~Qian, P.~Schwarz, {\em et~al.}, ``The {Kaldi}
  speech recognition toolkit,'' in {\em Proc. ASRU}, (Atlanta, USA), 2011.

\bibitem{rousseau2014enhancing}
A.~Rousseau, P.~Del{\'e}glise, and Y.~Est{\`e}ve, ``Enhancing the {TED-LIUM}
  corpus with selected data for language modeling and more {TED} talks,'' in
  {\em Proc. LREC}, (Reykjavik, Iceland), pp.~3935--3939, 2014.

\bibitem{hernandez2018ted}
F.~Hernandez, V.~Nguyen, S.~Ghannay, N.~Tomashenko, and Y.~Est{\`e}ve,
  ``{TED-LIUM 3:} twice as much data and corpus repartition for experiments on
  speaker adaptation,'' {\em arXiv preprint arXiv:1805.04699}, 2018.

\bibitem{panayotov2015librispeech}
V.~Panayotov, G.~Chen, D.~Povey, and S.~Khudanpur, ``Librispeech: an {ASR}
  corpus based on public domain audio books,'' in {\em Proc. ICASSP},
  (Brisbane, Australia), pp.~5206--5210, 2015.

\bibitem{han2017capio}
K.~J. Han, A.~Chandrashekaran, J.~Kim, and I.~Lane, ``The {CAPIO} 2017
  conversational speech recognition system,'' {\em arXiv preprint
  arXiv:1801.00059}, 2017.

\bibitem{garofolo1993darpa}
J.~S. Garofolo, L.~F. Lamel, W.~M. Fisher, J.~G. Fiscus, and D.~S. Pallett,
  ``{DARPA} {TIMIT} acoustic-phonetic continous speech corpus {CD-ROM}. {NIST}
  speech disc {1-1.1},'' {\em {NASA STI/Recon technical report n}}, vol.~93,
  1993.

\bibitem{godfrey1992switchboard}
J.~J. Godfrey, E.~C. Holliman, and J.~McDaniel, ``{SWITCHBOARD}: Telephone
  speech corpus for research and development,'' in {\em Proc. ICASSP}, (San
  Francisco, CA, USA), pp.~517--520, 1992.

\bibitem{cieri2004fisher}
C.~Cieri, D.~Miller, and K.~Walker, ``The {Fisher} corpus: a resource for the
  next generations of speech-to-text.,'' in {\em LREC}, vol.~4, (Lisbon,
  Portugal), pp.~69--71, 2004.

\bibitem{saon2017english}
G.~Saon, G.~Kurata, T.~Sercu, K.~Audhkhasi, S.~Thomas, D.~Dimitriadis, X.~Cui,
  B.~Ramabhadran, M.~Picheny, L.-L. Lim, B.~Roomi, and P.~Hall, ``English
  conversational telephone speech recognition by humans and machines,'' in {\em
  Proc. Interspeech 2017}, (Stockholm, Sweden), pp.~132--136, 2017.

\bibitem{oualil2017context}
Y.~Oualil, D.~Klakow, G.~Szasz{\'a}k, A.~Srinivasamurthy, H.~Helmke, and
  P.~Motlicek, ``A context-aware speech recognition and understanding system
  for air traffic control domain,'' in {\em Proc. ASRU}, (Okinawa, Japan),
  pp.~404--408, 2017.

\bibitem{milde2016ambient}
B.~Milde, J.~Wacker, S.~Radomski, M.~M{\"u}hlh{\"a}user, and C.~Biemann,
  ``Ambient search: A document retrieval system for speech streams,'' in {\em
  Proc. COLING 2016}, (Osaka, Japan), pp.~2082--2091, 2016.

\bibitem{Baumann2018}
T.~Baumann, A.~K{\"o}hn, and F.~Hennig, ``The spoken {Wikipedia} corpus
  collection: Harvesting, alignment and an application to hyperlistening,''
  {\em Language Resources and Evaluation}, Jan 2018.

\bibitem{sphinx2003}
P.~Lamere, P.~Kwok, W.~Walker, E.~Gouvea, R.~Singh, and P.~Wolf, ``Design of
  the {CMU} {S}phinx-4 decoder,'' in {\em Proceedings of Eurospeech}, (Geneva,
  Switzerland), pp.~1181--1184, 2003.

\bibitem{radeck2015open}
S.~Radeck-Arneth, B.~Milde, A.~Lange, E.~Gouv{\^e}a, S.~Radomski,
  M.~M{\"u}hlh{\"a}user, and C.~Biemann, ``Open source german distant speech
  recognition: Corpus and acoustic model,'' in {\em Proc. Text, Speech, and
  Dialogue (TSD)}, (Pilsen, Czech Republic), pp.~480--488, 2015.

\bibitem{schroder2003german}
M.~Schr{\"o}der and J.~Trouvain, ``The german text-to-speech synthesis system
  {MARY}: A tool for research, development and teaching,'' {\em International
  Journal of Speech Technology}, vol.~6, no.~4, pp.~365--377, 2003.

\bibitem{BASSAMPA}
{Bavarian Archive for Speech Signals}, ``Extended sam-pa.''
  \url{http://www.bas.uni-muenchen.de/forschung/Bas/BasSAMPA}.

\bibitem{bisani2008joint}
M.~Bisani and H.~Ney, ``Joint-sequence models for grapheme-to-phoneme
  conversion,'' {\em Speech communication}, vol.~50, no.~5, pp.~434--451, 2008.

\bibitem{kneser1995improved}
R.~Kneser and H.~Ney, ``Improved backing-off for m-gram language modeling,'' in
  {\em Proc. ICASSP}, (Detroit, MI, USA), pp.~181--184, 1995.

\bibitem{xu2018neural}
H.~Xu, K.~Li, Y.~Wang, J.~Wang, S.~Kang, X.~Chen, D.~Povey, and S.~Khudanpur,
  ``Neural network language modeling with letter-based features and importance
  sampling,'' in {\em Proc. ICASSP}, (Calgary, Alberta, Canada), 2018.

\bibitem{waibel1990phoneme}
A.~Waibel, T.~Hanazawa, G.~Hinton, K.~Shikano, and K.~J. Lang, ``Phoneme
  recognition using time-delay neural networks,'' in {\em Readings in speech
  recognition}, pp.~393--404, 1990.

\bibitem{peddinti2015time}
V.~Peddinti, D.~Povey, and S.~Khudanpur, ``A time delay neural network
  architecture for efficient modeling of long temporal contexts,'' in {\em
  Proc. Interspeech}, (Dresden, Germany), 2015.

\bibitem{saon2013speaker}
G.~Saon, H.~Soltau, D.~Nahamoo, and M.~Picheny, ``Speaker adaptation of neural
  network acoustic models using i-vectors.,'' in {\em Proc. ASRU}, (Olomouc,
  Czech Republic), pp.~55--59, 2013.

\bibitem{senior2014improving}
A.~Senior and I.~Lopez-Moreno, ``Improving dnn speaker independence with
  i-vector inputs,'' in {\em Acoustics, Speech and Signal Processing (ICASSP),
  2014 IEEE International Conference on}, (Florence, Italy), pp.~225--229,
  2014.

\bibitem{miao2015speaker}
Y.~Miao, H.~Zhang, and F.~Metze, ``Speaker adaptive training of deep neural
  network acoustic models using i-vectors,'' {\em IEEE/ACM Transactions on
  Audio, Speech and Language Processing (TASLP)}, vol.~23, no.~11,
  pp.~1938--1949, 2015.

\bibitem{povey2016purely}
D.~Povey, V.~Peddinti, D.~Galvez, P.~Ghahremani, V.~Manohar, X.~Na, Y.~Wang,
  and S.~Khudanpur, ``Purely sequence-trained neural networks for {ASR} based
  on lattice-free {MMI},'' in {\em Proc. Interspeech}, (San Fransisco, USA),
  pp.~2751--2755, 2016.

\bibitem{schnelle2014open}
D.~Schnelle-Walka, S.~Radeck-Arneth, C.~Biemann, and S.~Radomski, ``An open
  source corpus and recording software for distant speech recognition with the
  microsoft kinect,'' in {\em Proc. ITG}, (Erlangen, Germany), 2014.

\bibitem{wahlster2013verbmobil}
W.~Wahlster, {\em Verbmobil: foundations of speech-to-speech translation}.
\newblock Springer-Verlag Berlin/Heidelberg, 2000.

\bibitem{gaida2014comparing}
C.~Gaida, P.~Lange, R.~Petrick, P.~Proba, A.~Malatawy, and D.~Suendermann-Oeft,
  ``Comparing open-source speech recognition toolkits,'' {\em Tech. Rep., DHBW
  Stuttgart}, 2014.

\bibitem{smit2017improved}
P.~Smit, S.~Virpioja, M.~Kurimo, {\em et~al.}, ``Improved subword modeling for
  {WFST}-based speech recognition,'' in {\em Proc. Interspeech}, (Stockholm,
  Sweden), pp.~2551--2555, 2017.

\end{thebibliography}

\end{document}